\documentclass{article}

\PassOptionsToPackage{square,numbers}{natbib}
\usepackage[preprint]{neurips_2024}

\usepackage[utf8]{inputenc} 
\usepackage[T1]{fontenc}    
\usepackage{hyperref}       
\usepackage{url}            
\usepackage{booktabs}       
\usepackage{amsfonts}       
\usepackage{nicefrac}       
\usepackage{microtype}      
\usepackage{xcolor}         
\usepackage{graphicx}
\usepackage{color,soul}
\usepackage{lipsum}  
\usepackage{amsmath}
\usepackage{multirow}
\usepackage{colortbl}

\usepackage{todonotes} 
\usepackage{cleveref}
\usepackage{arydshln}
\usepackage{afterpage}

\def\cocacolaemoji{\raisebox{-0.3em}{\includegraphics[height=1.6em]{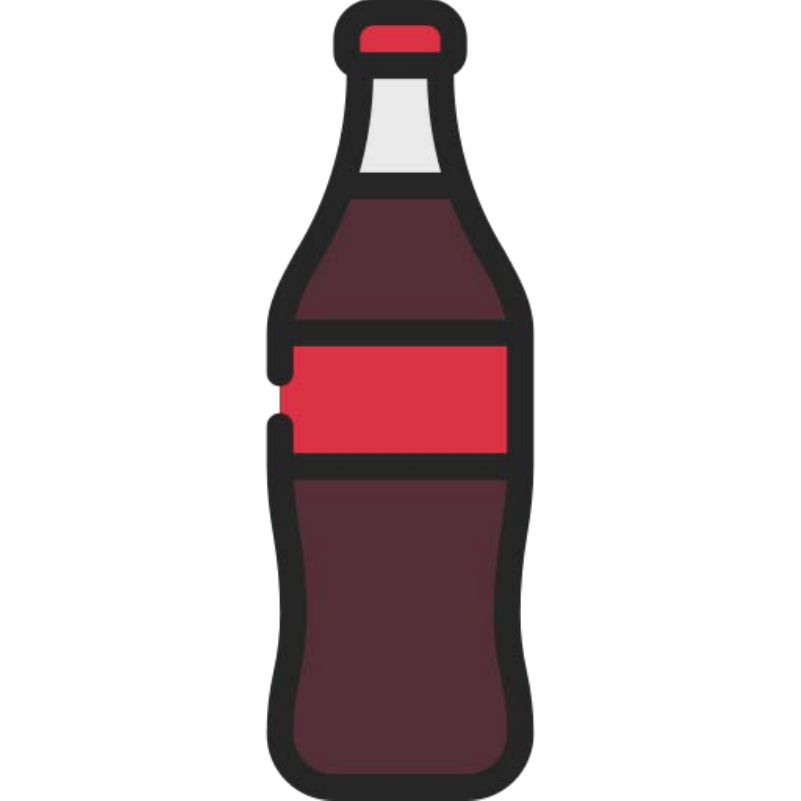}}}

\title{\cocacolaemoji CoCo-CoLa: Evaluating and Improving \\ Language Adherence in Multilingual LLMs}

\author{%
  Elnaz Rahmati\thanks{Equal contribution.} \And Alireza S. Ziabari\footnotemark[1] \And Morteza Dehghani \AND
        \normalfont \small University of Southern California \\
        \texttt{\small\{erahmati, salkhord, mdehghan\}@usc.edu}
}

\begin{document}

\maketitle

\begin{abstract}
  Multilingual Large Language Models (LLMs) develop cross-lingual abilities despite being trained on limited parallel data. However, they often struggle to generate responses in the intended language, favoring high-resource languages such as English. In this work, we introduce \emph{CoCo-CoLa} (Correct Concept - Correct Language), a novel metric to evaluate language adherence in multilingual LLMs. Using fine-tuning experiments on a closed-book QA task across seven languages, we analyze how training in one language affects others' performance. Our findings reveal that multilingual models share task knowledge across languages but exhibit biases in the selection of output language. We identify language-specific layers, showing that final layers play a crucial role in determining output language. Accordingly, we propose a partial training strategy that selectively fine-tunes key layers, improving language adherence while  reducing computational cost. Our method achieves comparable or superior performance to full fine-tuning, particularly for low-resource languages, offering a more efficient multilingual adaptation.\footnote{Our code is available at \href{https://github.com/elnazrahmati/CoCo-CoLa/}{https://github.com/elnaz rahmati/CoCo-CoLa/}}
\end{abstract}

\section{Introduction}

Multilingual LLMs are pre-trained on raw text from multiple languages, typically consisting of separate corpora for each language. Remarkably, despite this lack of explicit parallel data to facilitate cross-lingual associations, these models develop an implicit understanding of inter-language relations and cross-lingual word associations  \citep{wen-yi-mimno-2023-hyperpolyglot}. Instruction tuning further enhances their ability to follow prompts, and models trained on multilingual data often exhibit zero-shot cross-lingual transfer of instruction-following capabilities \citep{chirkova-nikoulina-2024-zero-shot}. However, this generalization is uneven: while high-resource languages in pretraining benefit significantly from instruction tuning, lower-resource or unseen languages often struggle to follow instructions reliably, frequently exhibiting degraded performance or defaulting to generating output in a preferred language \citep{nguyen-etal-2024-democratizing,chirkova-nikoulina-2024-zero-shot}.  To address these issues, we investigate how multilingual LLMs learn the same task across different languages.  

\begin{figure}
    \centering
    \includegraphics[width=0.8\linewidth]{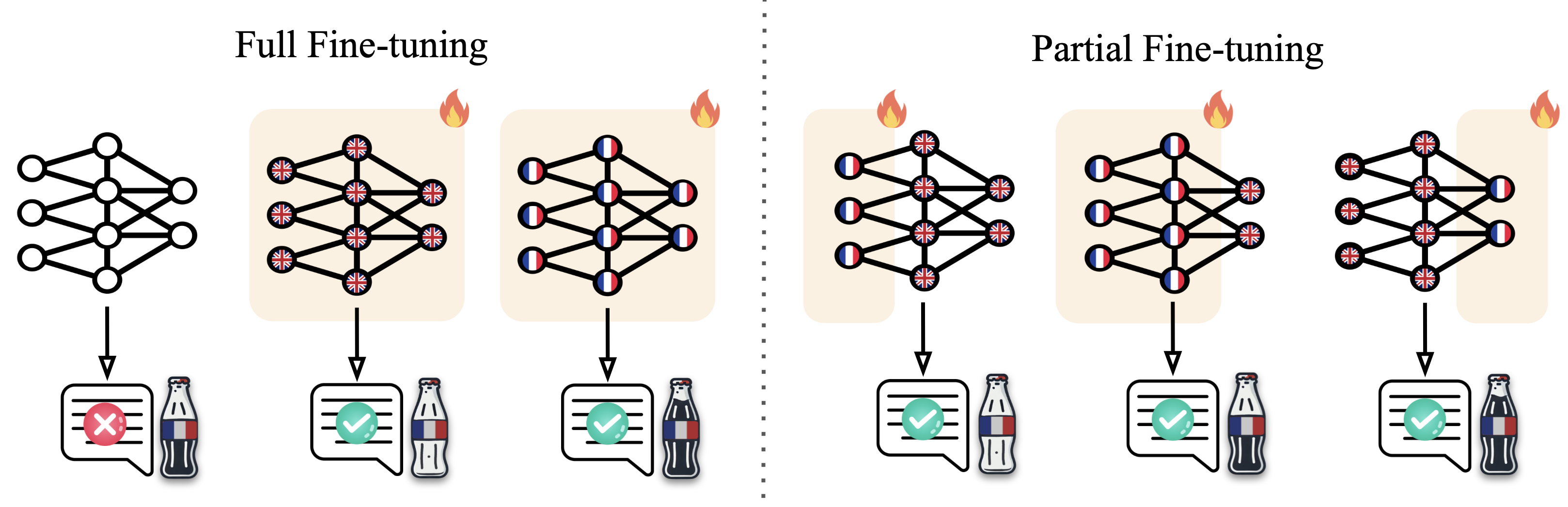}
    \caption{Evaluation of correctness and language adherence on French input. The soda level visualizes the CoCo-CoLa ratio, with higher levels indicating stronger adherence to the input language. Our results show that partially fine-tuning the final layers of an English-tuned model on French achieves language adherence and accuracy comparable to a model fully fine-tuned on French.}
    \label{fig:enter-label}
\end{figure}

A crucial step toward addressing the limitations of multilingual LLMs is understanding how they internally process and encode multilingual knowledge. Interpretability research has traditionally focused on monolingual models, leveraging techniques such as representation probing \citep{orgad2024llms,saphra-lopez-2019-understanding} and model patching \citep{ghandeharioun2024patchscope,garcia2024extracting}. These methods have been widely used to examine LLMs’ performance across tasks such as mathematics \citep{nikankin2024arithmetic,zhou2024pre}, and general knowledge \citep{jiang-etal-2024-large,burns2022discovering,singh2024rethinking,golgoon2024mechanistic,chowdhury2024probing,rai2024practical}. Studies on model internals suggest that Multi-Layer Perceptrons (MLPs) retrieve task-relevant information, while attention layers refine and promote the correct response \citep{geva-etal-2021-transformer, meng2022locating}. Furthermore, knowledge is often identified in earlier layers and reinforced in later layers \citep{fan2024not}.

However, these interpretability techniques have primarily been applied to monolingual models, which were initially dominant due to the early focus on English-language pertaining \cite{touvron2023llama,jiang2023mistral,team2024gemma,abdin2024phi}. The rise of multilingual LLMs trained on diverse languages \citep{gao-etal-2024-multilingual, shaham-etal-2024-multilingual, soykan2024linguistically}, necessitates extending interpretability research beyond English. Multilingual LLMs present additional challenges: representations of different languages are intertwined within a shared space; cross-lingual alignment varies across languages; and shared tokens between languages impact their process. These complexities make it difficult to isolate language-specific knowledge, benchmark cross-lingual generalization, and interpret how multilingual LLMs acquire and apply linguistic information. Given the prevalence of mid- and low-resource languages, understanding these mechanisms is crucial not only for improving cross-lingual transfer but also for mitigating the ``curse of multilinguality'' — the performance degradation observed as the number of supported languages increases.  

Recent efforts have begun addressing these challenges by probing internal multilingual representations \citep{li2024exploring}, analyzing the emergence of cross-lingual transfer \citep{wang-etal-2024-probing-emergence}, and studying token representation alignment on cross-lingual transfer \citep{gaschi-etal-2023-exploring}. Furthermore, researchers attempt to separate the linguistic abilities from task abilities by developing language- and task-specific adapters \citep{pfeiffer-etal-2020-mad,parovic-etal-2023-cross}, subnetworks \citep{choenni-etal-2023-cross}, or layers \citep{bandarkar2024layer}. However, despite this progress, most prior works treat multilinguality as a monolithic phenomenon, focusing on general cross-lingual transfer or aggregating all languages into a single block of linguistic knowledge. Less attention has been given to understanding how LLMs process individual languages at a more granular level, particularly within the context of task learning.

In this work, we focus on language adherence by first
identifying both shared and distinct patterns in cross-lingual task acquisition, revealing how multilingual models internalize and apply linguistic knowledge (\Cref{sec:preliminary}). We find that training on a task in one language improves performance in other languages. However, this benefit is not always directly observable due to an inherent model bias towards generating output in a preferred language, rather than strictly adhering to the input language (\Cref{sec:coca-cola-intro}). To quantify this bias, we introduce \emph{CoCo-CoLa} (Correct Concept, Correct Language), a novel metric designed to assess a model’s ability to generate responses in the intended input language, particularly for languages not included in supervised finetuning (SFT). Furthermore, we propose a \emph{partial training method} that selectively fine-tunes specific model layers which reveals the relation between language adherence and model layers
(\Cref{sec:partial-training}). This approach enables more efficient language adaptation, achieving comparable or even superior performance compared to 
full model retraining, especially for low-resource languages. Finally, we show that the issue of language adherence can be addressed by finetuning only the final layers of LLMs on a small balanced multilingual data (\Cref{sec:multi-training}).

\section{Related Work}

This work builds on several active research areas that inform our study of multilingual task learning in LLMs. Specifically, we draw from (1) Multilingual interpretability, which helps us analyze how LLMs process different languages and how their internal structures influence multilingual task learning; (2) Representation alignment, which provides insights into token-level similarities across languages and how shared representations facilitate cross-lingual generalization; (3) Adapters, which separate language knowledge from task-specific knowledge, offering a structured framework for understanding their interactions; and (4) Subnetworks, which identify task- and language-specific parameters within existing models, offering an alternative to external adapters and directly informing our approach to efficient partial training.

\paragraph{Interpretability.}
\citet{li2024exploring} use probing techniques to analyze accuracy changes across layers in LLMs, showing that high-resource languages exhibit patterns similar to English, with accuracy increasing from lower to upper layers. However, this pattern is inconsistent for low-resource languages. \citet{wang2024probing} examine cross-lingual transfer by analyzing neuron overlap in different languages using checkpoints from BLOOM’s pre-training \citep{le2023bloom}. They find a strong correlation between neuron overlap and cross-lingual transfer, though neuron overlap does not increase monotonically during training, and patterns vary across model sizes. Similarly, \citet{zhao2024large} investigate language-specific neurons and assess how  these neurons affect both English and non-English language performance.   

\paragraph{Representation alignment.}

Beyond studying multilingualism in LLMs, some research focuses on improving model performance across languages through representation alignment. \citet{gaschi-etal-2023-exploring} align English and Arabic model representations using a bilingual dictionary before fine-tuning on a target task. \citet{zhang-etal-2024-getting} align English representations with other languages using question-translation data before instruction-tuning. Additionally, \citet{salesky-etal-2023-multilingual} introduce a pixel representation method to enhance alignment and improve translation quality.   

\paragraph{Adapters.}

Another approach for cross-lingual transfer involves integrating adapters into the model. This technique is based on the assumption that task-solving knowledge can be separated from language knowledge. \citet{pfeiffer-etal-2020-mad} introduce MAD-X, a framework where language and task adapters are trained separately, with each block’s representations passing through a language adapter before a task adapter. Building on this, later works aim to refine adapter creation and composition methods. For instance, \citet{parovic-etal-2022-bad} propose BAD-X, which replaces monolingual adapters with bilingual adapters, improving performance for low-resource languages. \citet{zhao2024adamergex} introduce AdaMergeX, where adapters for language-task pairs are trained independently and later combined through linear operations (addition and subtraction) to generate adapters for new language-task pairs.

\paragraph{Subnetworks.}

To enhance cross-lingual transfer without adding new parameters, some methods focus on identifying existing task- and language-specific parameters within the model. \citet{choenni-etal-2023-cross} fine-tune models for specific languages or tasks, extract the most affected neurons, and use the resulting subnetworks to enable multilingual task adaptation. \citet{bandarkar2024layer} take a layer-wise approach in multiple steps: they train separate language- and task-expert models, analyze parameter changes to identify key layers for language and task learning, and use layer-swapping techniques to create a math expert in a new language. Consistent with \citet{zhao2024large}, their findings suggest that initial and final layers primarily encode linguistic information, while middle layers are task-specific.

\section{Preliminary Analysis} \label{sec:preliminary}

In the preliminary section of this paper, we first isolate language effects from task learning by choosing multi-lingual parallel  QA data (\Cref{sec:preliminary-setup}), examine fine-tuning performance across multiple languages (\Cref{sec:preliminary-performance}), explore how well LLMs generalize knowledge across languages (\Cref{sec:preliminary-shared-knowledge}), and which model components are most affected during training (\Cref{sec:parameter_update}). Then, in \Cref{sec:coca-cola-intro}, we introduce \textbf{CoCo-CoLa} metric to measure language adherence in multilingual LLMs followed by an efficient partial training method to increase the model adherence (\Cref{sec:partial-training}). 

\subsection{Setup} \label{sec:preliminary-setup}

To investigate how multilingual LLMs learn a new task in a monolingual setting, we train four different models on a Closed-Book Question-Answering (CBQA) task. We include two sizes of the Llama-3.2 series \citep{dubey2024llama} to analyze the effect of model size on multilingual performance and behavior, given that these models are specifically optimized for multilingual dialogue. We also include Llama-3.1-8B as a point of comparison, as it, while not explicitly optimized for multilingualism, was trained on a small multilingual corpus. To test generalizability to multilingualy balanced models, we include Gemma-3-4B \citep{gemmateam2025gemma3technicalreport}, which was trained with UniMax  \citep{chung2023unimaxfairereffectivelanguage} for addressing language imbalances.

We select CBQA because it is language-dependent and demonstrates a model's ability to act as a knowledge base \citep{wang-etal-2021-generative}. To isolate the impact of language differences from the effects of learning a new task or acquiring new knowledge, we use the Mintaka CBQA dataset \citep{sen-etal-2022-mintaka}.  Mintaka provides identical question-answer pairs in nine languages, allowing us to keep the question content consistent and thus isolate the influence of language itself. The dataset was originally created in English and later translated into Arabic, French, German, Hindi, Italian, Japanese, Portuguese, and Spanish.  

One challenge with Mintaka is that some answer types are not translated across languages. To keep question-answer pairs within the same language, we use Google Translate to convert these answers into the language of their respective questions and apply back-translation for accuracy checks. Additionally, since our goal is to study how models learn new tasks in languages they have been exposed to before, we exclude Arabic and Japanese.

\subsection{SFT Performance} \label{sec:preliminary-performance}
\begin{table}[t]
\centering
\caption{Performance of pre-trained models (PLM), fine-tuned models (SFT), and their difference ($\Delta$ = SFT - PLM) on CBQA data across languages.}
\label{tab:initial_acc}
\resizebox{0.95\columnwidth}{!}{ 
\begin{tabular}{lcccccccccccc}
\toprule
 & \multicolumn{3}{c}{\textbf{Llama-1B}} & \multicolumn{3}{c}{\textbf{Llama-3B}} & \multicolumn{3}{c}{\textbf{Llama-8B}} & \multicolumn{3}{c}{\textbf{Gemma-4B}} \\ 
\cmidrule(lr){2-4} \cmidrule(lr){5-7} \cmidrule(lr){8-10} \cmidrule(lr){11-13}
\textbf{Language} & PLM & SFT & $\Delta$ & PLM & SFT & $\Delta$ & PLM & SFT & $\Delta$ & PLM & SFT & $\Delta$ \\ 
\midrule
English     & 13.27 & 38.44 & 25.17 & 32.85 & 53.09 & 20.24 & 12.92 & 50.98 & 38.06 & 30.69 & 53.67 & 22.98 \\
French      & 11.30 & 40.27 & 28.97 & 22.90 & 43.80 & 20.90 & 18.53 & 50.85 & 32.32 & 21.67 & 48.43 & 26.76 \\
German      & 7.16  & 40.34 & 33.18 & 23.79 & 48.10 & 24.31 & 11.04 & 44.35 & 33.31 & 23.76 & 45.79 & 22.03 \\
Hindi       & 5.27  & 21.18 & 15.91 & 7.33  & 30.39 & 23.06 & 6.21  & 35.29 & 29.08 & 8.84 & 43.96 & 35.12 \\
Italian     & 7.06  & 41.58 & 34.52 & 21.87 & 42.73 & 20.86 & 16.48 & 43.22 & 26.74 & 20.44 & 50.05 & 29.61 \\
Portuguese  & 5.38  & 38.23 & 32.85 & 20.06 & 37.04 & 16.98 & 18.38 & 31.11 & 12.73 & 19.96 & 44.16 & 24.20 \\
Spanish     & 6.13  & 41.71 & 35.58 & 22.01 & 45.69 & 23.68 & 16.60 & 45.46 & 28.86 & 26.33 & 48.13 & 21.80 \\
\bottomrule
\end{tabular}
}

\end{table}

Our initial step is to assess the model's ability to learn the task in each individual language, effectively measuring how learning difficulty varies across languages. To do this, we perform SFT for all models on each language of the CBQA dataset for three epochs and generate answers for given questions. Next, we select the best model based on the validation loss. Further implementation details are provided in \Cref{sec:implementation-detail}.

\autoref{tab:initial_acc} shows a comparison of accuracy between the pre-trained model and the best checkpoint of the language-specific SFT model across different languages. SFT significantly improves performance for all languages with relatively consistent accuracy levels, except for Hindi in all Llama model sizes and Portuguese for Llama-8B, which exhibit notably lower accuracy. This discrepancy is likely due to undertraining. Among the SFT models, English achieves the highest accuracy in 
all models, except Llama-1B that performs best in Spanish.
The largest accuracy gains are observed in English (+38.06\%) for Llama-8B, German (+24.31\%) for Llama-3B, Spanish (+35.58\%) for Llama-1B, and Hindi (+35.12\%) for Gemma-4B,  indicating that these languages benefited the most from fine-tuning. The comparable accuracy across languages indicates comparable knowledge acquisition.  

However, two critical questions remain: (1) Do models share learned knowledge uniformly across languages, or do they correctly answer distinct subsets of questions depending on the language? (2) Are there specific parts of the model that are responsible for encoding language-specific information? To address these questions, we first analyze the overlap in correct answers across languages using the Jaccard Index, followed by an investigation of parameter updates to determine whether certain components of the model specialize in handling linguistic differences.

\begin{figure*}[t]
    \centering
    \includegraphics[width=\linewidth]{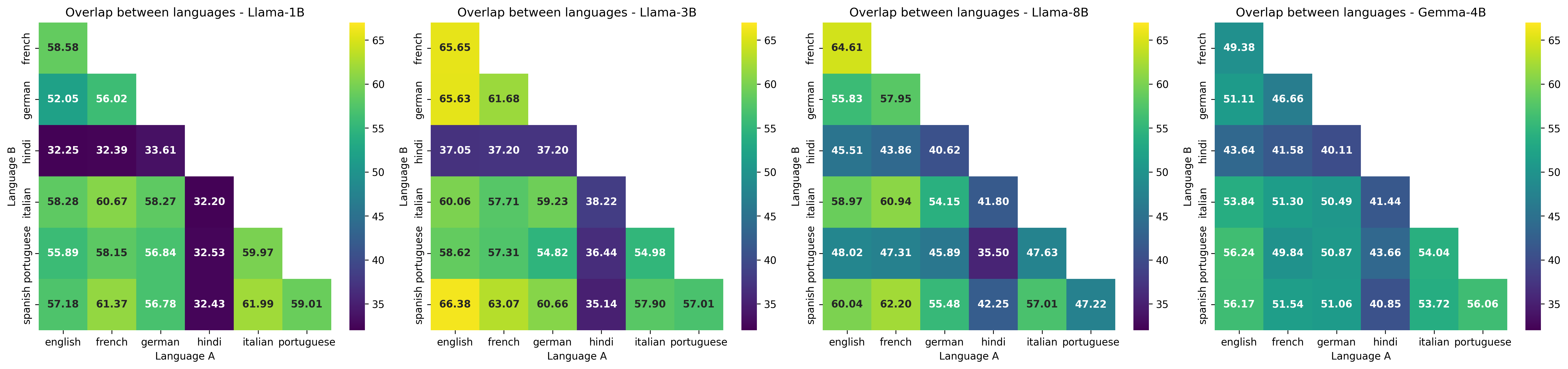}
    \caption{Jaccard similarity index between different languages, measuring the proportion of overlapping correctly answered questions between pairs of languages.}
    \label{fig:overlap}
\end{figure*}

\subsection{Cross-lingual Task Knowledge} \label{sec:preliminary-shared-knowledge}

To further investigate the extent of cross-lingual task knowledge transfer within the model, we analyze the overlap in correct answers across languages. Specifically, we measure how consistently the model arrives at the same correct answers in different languages, providing insight into whether knowledge is shared across languages. It is important to note that there is no overlap between the knowledge present in the training and evaluation data. This ensures that any correct answers during evaluation are derived from knowledge acquired during pretraining rather than memorization. Consequently, the model’s ability to generate correct responses across languages indicates that it has internalized the underlying task knowledge from the training data, rather than relying solely on language-specific cues. Let \( L_A \) and \( L_B \) represent two languages, and let \( C_{L_A} \) denote the set of correct answers for \( L_A \). To quantify the degree of shared task knowledge between languages, we compute the Jaccard Index, also known as Intersection over Union (IoU), between \( C_{L_A} \) and \( C_{L_B} \) (see \autoref{eq:overlap}). The Jaccard Index is a natural choice for this analysis as it directly measures the proportion of overlapping correct answers relative to the total distinct answers across languages. This allows us to assess knowledge consistency and cross-lingual transfer within the model.
\vspace{-1mm}
\begin{equation}
    IoU(A, B) = \frac{|C_{L_A} \cap C_{L_B}|}{|C_{L_A} \cup C_{L_B}|}
    \label{eq:overlap}
\end{equation}

The results, shown in \autoref{fig:overlap}, indicate that on average approximately 60\% of correctly answered questions are shared across languages for all models, suggesting a strong degree of shared knowledge among languages. However, Hindi exhibits significantly lower overlap with other languages in Llama-3.2 models, suggesting weaker generalization for this language. Interestingly, in Llama-8B, Hindi shows higher overlap compared to Llama-3.2 models, but Portuguese experiences a notable drop in overlap. Additionally, Llama-3B demonstrates a higher rate of shared knowledge compared to Llama-8B, despite both models achieving comparable accuracy across languages (see \autoref{tab:initial_acc}). This highlights the importance of multilingual optimization in enhancing cross-lingual transfer among languages. For Gemma-4B, despite comparable accuracy across languages, the overall overlap is lower than that observed in the Llama models, indicating less cross-lingual knowledge sharing.

\subsection{Parameter Updates} \label{sec:parameter_update}

To investigate language-specific encoding in LLMs, we analyze parameter updates during fine-tuning and compare them across languages to determine whether certain components of the model specialize in processing linguistic information.  \citet{meng2022locating} suggest that MLP modules primarily store knowledge, while attention modules control information retrieval and selection. SFT models correctly answer approximately 40\% of evaluation questions in all languages. However, they require fine-tuning to improve their ability to select and output the correct information. As a result, we expect substantial modifications in the attention modules, particularly in the final layers, while changes in the MLP modules remain limited. Since these datasets differ only in language, not in task or knowledge, analyzing the model updates allows us to pinpoint which layers or components are most crucial for learning language-specific representations. 

To compute parameter update,  we follow \citet{bandarkar2024layer} and calculate the average parameter modifications for each module in each layer. Denoting the pre-trained weight matrix as \( W_p \) and the fine-tuned weight matrix as \( W_f \), the average magnitude of differences is given by: 
\begin{equation}
\Delta W = \frac{1}{n} \sum_{i=1}^{n} | W_p^{(i)} - W_f^{(i)} |
\label{eq:model-diff}
\end{equation}

\begin{figure*}
    \centering
    \includegraphics[width=0.9\linewidth]{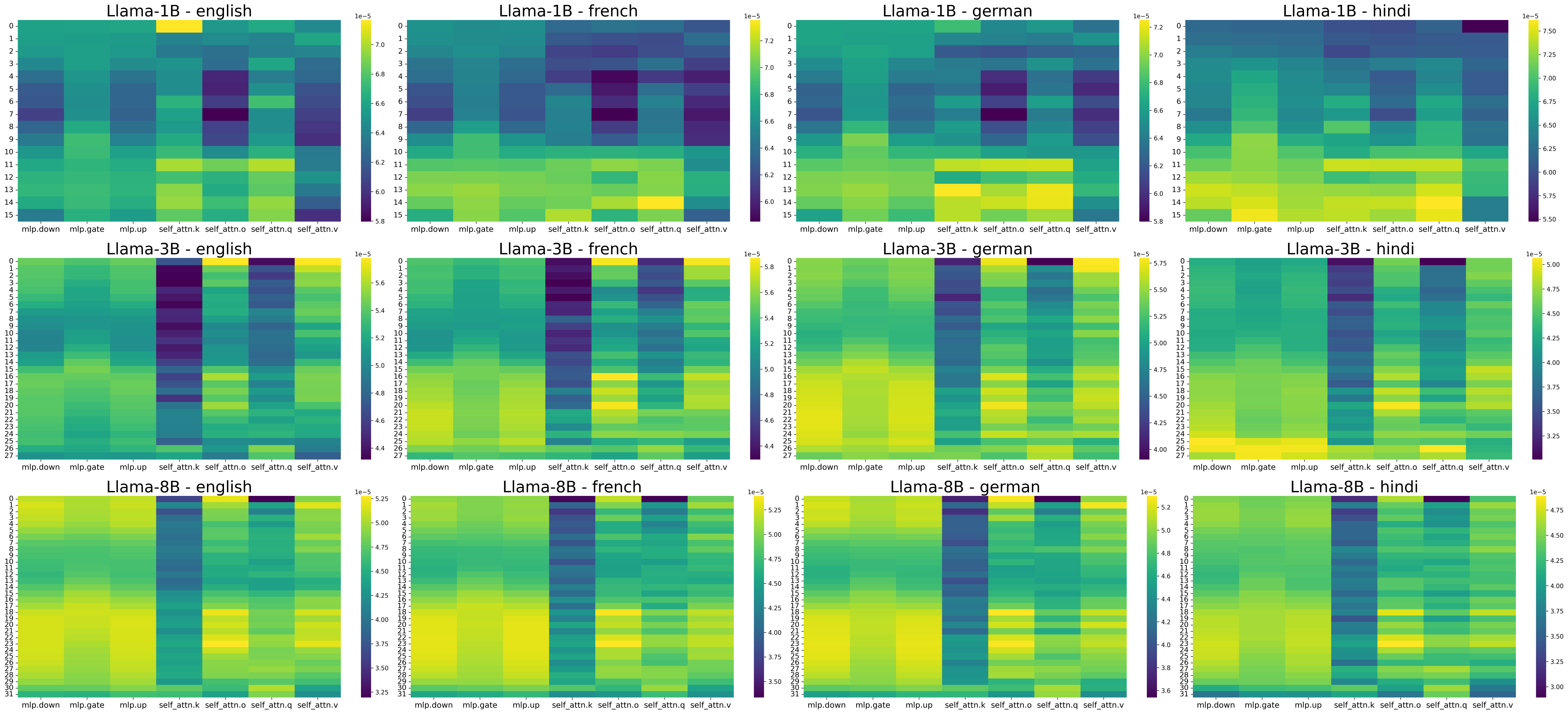}
    \caption{Heatmaps of parameter update magnitudes during monolingual fine-tuning on English (top) and French (bottom) across different LLMs.}
    \label{fig:model-diff}
\end{figure*}

The results for English and French are shown in \autoref{fig:model-diff}, with the remaining languages in \autoref{fig:model-diff-appendix}. As expected, significant modifications occur in the attention modules of the final six layers for Llama-1B and the final 14 layers for Llama-3B, Llama-8B, and Gemma-4B models across all languages. However, in Llama-3.2 models and Gemma-4B model, we observe substantial changes in the MLP modules in these layers for all languages except English, suggesting that these variations might be tied to language-specific processing rather than task-related learning. Surprisingly, for Llama-8B, even the model fine-tuned on English shows a high rate of change similar to other languages. Considering the unexpectedly low accuracy of the Llama-8B pre-trained model across all languages compared to Llama-3B, this larger modification could be related to learning the task or acquiring new knowledge rather than just language adaptation.

\section{Approach}

Our previous analysis suggests that while task knowledge is largely shared across languages, the \emph{way} this knowledge is processed and accessed differs. Although a Jaccard Index analysis revealed substantial overlap in correct answers, our investigation of parameter updates showed that models trained on non-English languages required more substantial modifications in their MLP modules compared to English, even when achieving comparable accuracy. This raises an important question: Do these modifications reflect deviations in knowledge acquisition, or are they more related to language generation? In this section, we first  introduce a metric to analyze linguistic bias in multilingual LLM outputs. Then, we propose a partial training strategy aimed at reducing this bias by selectively fine-tuning specific model components.

\subsection{Correct Concept in Correct Language} \label{sec:coca-cola-intro}

According to \citet{dubey2024llama}, only 8\% of the pre-training data used for Llama-3 models is multilingual, while the rest is dominated by English general knowledge, mathematics, and code. This suggests a strong bias toward English. Given this imbalance, we hypothesize that the observed MLP module changes in non-English languages may not indicate new knowledge acquisition but rather adjustment in language selection during response generation. Supporting this, \citet{chirkova-nikoulina-2024-zero-shot} found that when Llama-2-13B is instruction-tuned on English and tested in other languages, it generates responses in a different language from input language in over 30\% of cases, with this behavior influenced by training hyperparameters.  

To investigate this further, we introduce \textbf{CoCo-CoLa} (\textbf{Co}rrect \textbf{Co}ncept - \textbf{Co}rrect \textbf{La}nguage), a metric designed to measure how well the model adheres to the input language while generating correct responses. Let \( L_{i} \) denote the input language, \( C_{L_{i} \rightarrow L_{o}} \) the set of correct output in language \( L_{o} \) when passing language \( L_{i} \) as input. We define the CoCo-CoLa ratio as follows:  
\begin{equation}
    \text{CoCo-CoLa}(L_{i}) 
      = \frac{|C_{L_{i}\rightarrow L_{i}} - \bigcup\limits_{L_{o}\neq L_{i}}C_{L_{i}\rightarrow L_{o}}|}
    {|C_{L_{i}\rightarrow L_{i}} ~ \Delta  \bigcup\limits_{L_{o}\neq L_{i}}C_{L_{i}\rightarrow L_{o}}|}
    \label{eq:coca-cola}
\end{equation}

\begin{table*}[t]
\centering
\caption{CoCo-CoLa ratio (Ratio) and cumulative accuracy (Acc) of pretrained model (PLM), English-tuned model (\(\rightarrow en\)), and \(L_i\)-tuned model (\(\rightarrow L_i\)) across languages for Llama-1B, Llama-3B, Llama-8B, and Gemma-4B.}
\vspace{3mm}
\label{tab:model_performance}
\resizebox{\textwidth}{!}{ 
\begin{tabular}{ll|ccc|ccc|ccc|ccc}
\toprule
\multirow{2}{*}{\textbf{Language}} & \multirow{2}{*}{\textbf{Metric}} & \multicolumn{3}{c}{\textbf{Llama-1B}} & \multicolumn{3}{c}{\textbf{Llama-3B}} & \multicolumn{3}{c}{\textbf{Llama-8B}} & \multicolumn{3}{c}{\textbf{Gemma-4B}} \\ 
\cmidrule(lr){3-5} \cmidrule(lr){6-8} \cmidrule(lr){9-11} \cmidrule(lr){12-14}
& 
& $PLM$ & $\rightarrow en$  & $\rightarrow L_{i}$
& $PLM$ & $\rightarrow en$ & $\rightarrow L_{i}$
& $PLM$ & $\rightarrow en$ & $\rightarrow L_{i}$
& $PLM$ & $\rightarrow en$ & $\rightarrow L_{i}$ \\
\midrule
\multirow{2}{*}{French}      
& Acc   & 12.07 & 52.66 & 55.73 & 20.57 & 62.55 & 52.97 & 12.89 & 58.64  & 66.01 & 18.67 & 65.16 & 63.23 \\
& Ratio   & 49.42 & 13.47 & 88.58 & 52.51 & 14.73 & 89.45 & 58.11 & 12.32 & 87.54 & 50.77 & 19.22 & 90.14 \\ \hdashline
\multirow{2}{*}{German}      
& Acc   & 8.05  & 51.97 & 50.92 & 16.99 & 49.30 & 57.01 & 10.43 & 59.95 & 52.27 & 15.26 & 64.59 & 60.24 \\
& Ratio   & 53.87 & 10.50 & 91.02 & 56.53 & 19.64 & 89.26 & 57.49 & 11.03 & 87.21 & 42.82 & 15.23 & 92.64 \\\hdashline
\multirow{2}{*}{Hindi}      
& Acc   & 8.65  & 29.34 & 27.42 & 15.77 & 38.26 & 39.67 & 9.79  & 37.29 & 39.21 & 12.58 & 47.81 & 49.66 \\
& Ratio   & 43.16 & 13.28 & 90.79 & 31.93 & 10.04 & 77.47 & 43.67 & 10.74 & 90.68 & 40.86 & 9.39  & 97.19 \\\hdashline
\multirow{2}{*}{Italian}      
& Acc   & 7.76  & 51.35 & 62.39 & 16.63 & 53.17 & 46.02 & 11.77 & 61.88 & 58.55 & 14.62 & 62.99 & 67.98 \\
& Ratio   & 51.32 & 10.00 & 93.60 & 56.68 & 16.29 & 87.91 & 52.11 & 10.90 & 91.35 & 48.76 & 14.84 & 91.08 \\\hdashline
\multirow{2}{*}{Portuguese}      
& Acc   & 10.22 & 54.85 & 57.57 & 17.60 & 55.52 & 50.64 & 16.23 & 60.75 & 42.90 & 17.11 & 63.81 & 61.16 \\
& Ratio   & 56.40 & 12.73 & 91.07 & 63.37 & 15.99 & 85.10 & 51.41 & 11.49 & 90.73 & 51.89 & 14.98 & 90.69 \\\hdashline
\multirow{2}{*}{Spanish}      
& Acc   & 9.75  & 57.52 & 59.02 & 19.17 & 57.55 & 60.38 & 14.13 & 58.34 & 54.27 & 17.69 & 65.88 & 60.65 \\
& Ratio   & 52.28 & 12.01 & 91.24 & 61.68 & 15.84 & 89.18 & 61.98 & 9.40  & 91.35 & 51.15 & 14.70 & 91.36 \\
\bottomrule
\end{tabular}
}

\end{table*}
\vspace{-2mm}
The denominator uses the symmetric difference between \( C_{L_{i}\rightarrow L_{i}} \) and correct answers in other languages because many answers involve named entities, such as well-known places, books, and individuals. Since most of the languages in this work use similar scripts, named entities often appear in identical forms across multiple languages. This redundancy leads to overlap between \( C_{L_{i} \rightarrow L_{i}} \) and \( \bigcup_{L_{o}\neq L_{i}}C_{L_{i}\rightarrow L_{o}} \), which the symmetric difference helps mitigate by ensuring that shared named entities do not artificially inflate the metric.

Given that these models are primarily trained on English, when the input is in \( L_{i} \) the output is usually either \( L_{i} \) or English. Thus, \( \bigcup_{L_{o}\neq L_{i}}C_{L_{i}\rightarrow L_{o}} \) is largely dominated by \(C_{L_{i}\rightarrow en}\), meaning that most language switching occurs between the input language and English rather than other languages.

To further simplify the calculation, we filter the data to include only questions where the correct answers in \( L_{i} \) and English are different. Under this condition, \( C_{L_{i} \rightarrow L_{i}} \cap C_{L_{i} \rightarrow en} = \emptyset \), allowing the CoCo-CoLa ratio to reduce to:  

\vspace{-3mm}
\begin{equation}
    \text{CoCo-CoLa}(L_{i}) = \frac{|C_{L_{i}\rightarrow L_{i}}|}{|C_{L_{i}\rightarrow L_{i}}| + |C_{L_{i}\rightarrow en}|}
    \label{eq:coca-cola-simplified}
\end{equation}
\vspace{-2mm}

To evaluate language adherence and accuracy, we pass the input in $L_{i}$ to pre-trained, \textit{en-tuned}, and  $L_{i}$\textit{-tuned} models. We then compute the CoCo-CoLa ratio and the cumulative accuracy, defined as the proportion of correct answers either in \( L_{i} \) or English. The results, presented in \autoref{tab:model_performance}, show that while the \textit{en-tuned} models and the $L_{i}$\textit{-tuned} models achieve comparable cumulative accuracy on $L_{i}$ input, the CoCo-CoLa ratio is significantly lower for the \textit{en-tuned} model. This suggests that although the \textit{en-tuned} model can correctly process the question in $L_{i}$ and retrieve the correct answer at the same rate as the $L_{i}$\textit{-tuned} model, it frequently generates the answer in English instead of \( L_{i} \). Furthermore, analyzing the CoCo-CoLa ratio of the pre-trained model reveals that the model already exhibits a bias toward generating English responses, though this bias is less pronounced than in the \textit{en-tuned} model. These findings support our hypothesis that the varying rate of parameter updates across languages is related to output language preference. Since the model is already inherently biased toward English, \textit{en-tuned} results in the least MLP change compared to other languages.

\subsection{Partial Training for Language Adaptation} \label{sec:partial-training}

In this section, we aim to disentangle task learning from output generation in language $L_{i}$. Our previous results reveal two key observations. First, as shown in \Cref{sec:coca-cola-intro}, both the \textit{en-tuned} model and the $L_{i}$\textit{-tuned} model achieve comparable cumulative accuracy on $L_{i}$, indicating that they learn the task equally well. The only difference is their CoCo-CoLa score, meaning that while both models understand the task to the same degree, they generate outputs in different languages. Second, from \Cref{sec:parameter_update}, we observed that the \textit{en-tuned} and  $L_{i}$\textit{-tuned} models undergo different parameter updates. Some of these updates are necessary for learning the task itself, while others may specifically steer the model toward producing responses in the intended language.
\begin{figure*}[ht]
    \centering
    \includegraphics[width=0.8\linewidth]{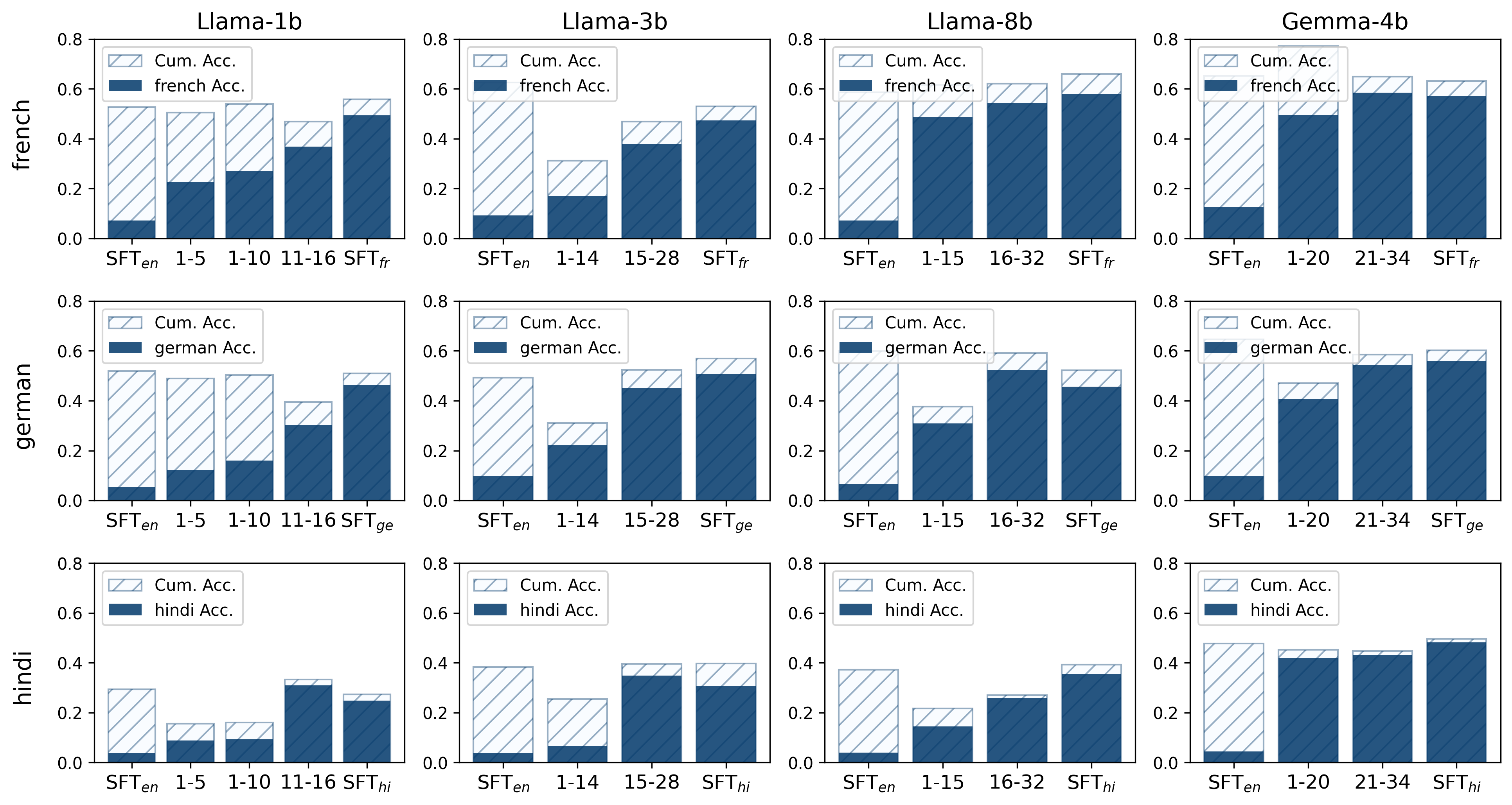}
    \caption{Cumulative accuracy and \( L_i \) accuracy on \textit{en-tuned} (\( SFT_{en} \)) and \( L_i \)-tuned models (\( SFT_{L_i} \)), along with partially trained models, across all Llama model sizes.}
    \label{fig:partial-training-full}
\end{figure*}

Based on these observations, we hypothesize that fine-tuning specific layers of an \textit{en-tuned} model on $L_{i}$ can enable it to generate responses in  $L_{i}$  without requiring full model updates. Specifically, these layers correspond to the parameters that were updated in the  $L_{i}$\textit{-tuned} model but not in the \textit{en-tuned} model. To test this hypothesis, we first identify the layers that undergo language-specific updates. We then fine-tune only these layers in the \textit{en-tuned} model and compare the results to fine-tuning other layers. This comparison allows us to isolate the parameters responsible for output language. 

\paragraph{Identifying language layers.} We select layers for partial training based on the variation in parameter update rates observed in \Cref{sec:parameter_update}. For the Llama-1B model, we train three variants by unfreezing different sets of layers: (1) layers 11-16, (2) layers 1-5 (chosen to match the parameter count of the final six layers), and layers 1-10 (including all parameters except the final six). We expect the first variant to be the most language-related and to result in the largest improvement in the CoCo-CoLa ratio, while the other two should have a smaller effect. For Llama-3B and Gemma-4B, we similarly train two variants each: unfreezing layers 15–28 and 1–14 for Llama-3B, and layers 21–34 and 1–20 for Gemma-4B. Again, we expect the final-layer variants to have a stronger relationship to language generation.
For Llama-8B, which does not show clear variations in update rates across languages (as noted in \Cref{sec:parameter_update}), we instead select layers based on the most updated MLP modules. Specifically, we choose layers 16–32 and layers 1–15 for partial training to determine which part of the model is more responsible for language generation. Through this analysis, we aim to verify whether the final layers play a greater role in controlling the output language

\paragraph{Partial training evaluation.} To evaluate the effectiveness of partial training, we compare all partially trained models to both their fully \textit{en-tuned} and fully \(L_{i}\)\textit{-tuned} models. \autoref{fig:partial-training-full} presents cumulative accuracy and \(L_{i}\) accuracy across three languages, while results for the remaining three languages are included in \autoref{fig:partial-training-appendix}. In addition, CoCo-CoLa ratios for partially trained models are also available in \Cref{sec:partial-appendix}, providing further insight into the extent to which partial fine-tuning improves output language consistency. 

As shown in \autoref{fig:partial-training-full}, among the partially trained models, unfreezing the final layers results in the highest accuracy and CoCo-CoLa ratio for all models, highlighting the crucial role these layers play in determining the output language. Notably, the accuracy of this partially trained configuration closely approaches that of the fully \(L_{i}\)\textit{-tuned} model, suggesting that the earlier layers already encode sufficient information for question answering, even without direct exposure to \(L_{i}\) during training.
Interestingly, Hindi-which initially exhibited lower performance than other languages-benefits significantly from cross-lingual transfer, achieving better results with partial training than with full training in both Llama-3.2 models. Llama-3B demonstrates even stronger cross-lingual transfer, with improved accuracy for Italian and Portuguese as well.
For Llama-8B and Gemma-4B, training the second half of the model yields the highest CoCo-CoLa ratio; however, the differences in \(L_{i}\) accuracy across partial training configurations are less pronounced than in the Llama-3.2 models. These models also show improved accuracy with partial training compared to full training for German, Italian, and Portuguese in Llama-8B, and for French, Portuguese, and Spanish in Gemma-4B.  

For low-resource languages, partially training only the final layers of an \textit{en-tuned} model can achieve similar or even better accuracy compared to full fine-tuning in the target language. Beyond its effectiveness, partial training is significantly more efficient, reducing training time to half and memory usage to 65\% of full training. Furthermore, the model achieves higher accuracy in fewer training steps, requiring less than one epoch, meaning it is trained on fewer data points.  

These findings confirm the hypothesis that the final layers are linked to output language selection, whereas initial and middle layers have less effect on the output language. Our results are aligned with concurrent work suggesting that LLMs process input in three stages: understanding the input, reasoning and knowledge retrieval in a shared space among languages, and generating output \citep{wendler-etal-2024-llamas,dumas2025separatingtonguethoughtactivation,schut2025multilingualllmsthinkenglish}. Although it remains debated whether this shared knowledge space is language agnostic \citep{dumas2025separatingtonguethoughtactivation} or whether the model simply thinks in English \citep{wendler-etal-2024-llamas,schut2025multilingualllmsthinkenglish}, these works, alongside ours, all suggest that the process happening in middle layers is not dependent on the input language. However, what previous works overlooks is that the final stage is defective and cannot generate the response in the correct language. We believe this phenomenon has led to misleading evaluations and the belief that multilingual LLMs think better in English \citep{etxaniz-etal-2024-multilingual}. Our work emphasizes the importance of considering both correctness and language adherence, as relying on output accuracy against the ground truth does not provide a complete picture of a model's ability to reason and operate in non-dominant languages.

\subsection{Improving Language Adherence in Multilingual LLMs} \label{sec:multi-training}
\begin{table}[]
    \centering
    \caption{CoCo-CoLa ratio (Ratio) and cumulative accuracy (Acc) of models partially trained on balanced multilingual data, with averages across all languages.}
        \label{tab:multi-training}
    \resizebox{\textwidth}{!}{ 
    \begin{tabular}{lcccccccccccccc}
    \toprule
     & \multicolumn{2}{c}{\textbf{French}} & \multicolumn{2}{c}{\textbf{German}} & \multicolumn{2}{c}{\textbf{Hindi}} & \multicolumn{2}{c}{\textbf{Italian}} & \multicolumn{2}{c}{\textbf{Portuguese}} & \multicolumn{2}{c}{\textbf{Spanish}} & \multicolumn{2}{c}{\textbf{Average}} \\

        \cmidrule(lr){2-3} \cmidrule(lr){4-5} \cmidrule(lr){6-7} \cmidrule(lr){8-9} \cmidrule(lr){10-11} \cmidrule(lr){12-13} \cmidrule(lr){14-15} 
     
     \textbf{Model} & Ratio & Acc & Ratio & Acc & Ratio & Acc & Ratio & Acc & Ratio & Acc & Ratio & Acc & Ratio & Acc \\
    \midrule

Llama-1B & 82.15 & 47.53 & 64.47 & 39.02 & 90.75 & 28.00 & 83.58 & 50.39 & 73.89 & 41.51 & 72.72 & 41.65 & 77.93 & 41.35 \\
Llama-3B & 78.37 & 42.19 & 79.55 & 36.14 & 85.34 & 33.54 & 80.11 & 41.61 & 74.27 & 49.04 & 78.92 & 44.22 & 79.43 & 41.12 \\
Llama-8B & 75.95 & 67.62 & 85.29 & 49.75 & 96.89 & 32.00 & 87.83 & 59.77 & 88.63 & 64.55 & 86.94 & 38.01 & 86.92 & 51.95 \\
Gemma-4B & 88.35 & 57.38 & 88.04 & 55.88 & 83.71 & 37.32 & 90.22 & 64.21 & 87.00 & 60.69 & 87.34 & 61.81 & 87.45 & 56.22 \\

    \bottomrule
    \end{tabular}
    }
\end{table}

As demonstrated in \Cref{sec:coca-cola-intro}, multilingual LLMs exhibit a strong linguistic bias toward English, the most prevalent language in their training data. In \Cref{sec:partial-training}, we further established that this bias is closely linked to the model’s final layers. To investigate whether this bias can be mitigated and to enable the model to better adhere to the input language, we take the \textit{en-tuned} model and, rather than adapting it to a single target language, we partially fine-tune the language-related layers using a balanced multilingual dataset, where all languages appear with equal frequency in the training data.  

As shown in \autoref{tab:multi-training}, the average CoCo-CoLa ratio for multilingually fine-tuned Gemma-4B and Llama-8B reaches 87.45\% and 86.92\%, respectively, while Llama-1B and Llama-3B achieve slightly lower ratios of 77.93\% and 79.43\%. 
These results are similar to the monolingual models partially trained for each language (\Cref{sec:partial-appendix}). These findings indicate that, even when starting from a model pretrained on biased data, fine-tuning only the final layers on a balanced multilingual dataset substantially improves language adherence across all languages. Notably, for Llama-8B and Gemma-4B, the accuracy of the resulting multilingual model is competitive with models fully fine-tuned for each individual language, despite using only 200 datapoints per language during training.

\section{Conclusion}
In this work, we first analyzed shared knowledge across seven languages and identified key differences in the parameters most affected when training models for each language. Building on these insights, we proposed the CoCo-CoLa ratio, a metric for evaluating language adherence in multilingual LLMs, and used it to evaluate both pre-trained and fine-tuned LLMs. Our findings show that pre-trained models tend to generate English outputs regardless of the input language and that fine-tuning on English further amplifies this bias.

To address this problem, we leveraged insights from parameter updates and CoCo-CoLa results to develop a partial training method that improves language adherence in English-trained models. Our analysis demonstrated a more efficient alternative to full fine-tuning, achieving comparable or even superior performance while significantly reducing the number of updated parameters. Additionally, we showed that partial training on balanced multilingual data achieves similar language adherence to monolingual training.
Given the widespread availability of instruction-tuned and task-specific English models, partial training of final layers presents a fast and efficient approach for improving language adherence and adapting LLMs to new languages.

\bibliography{custom}

\appendix
\section{Appendix}
\label{sec:appendix}

\subsection{Limitations}
We acknowledge that training hyperparameters can influence the linguistic bias of fine-tuned models, as highlighted by \citet{chirkova-nikoulina-2024-zero-shot}. For instance, while smaller learning rates may reduce bias, they can also lead to degraded task performance. Due to resource constraints, we used a single set of hyperparameters optimized for task performance. Additionally, we applied the same hyperparameter settings across all languages and model sizes, though fine-tuning them individually for each model-language pair could potentially yield better results.

Moreover, linguistic bias in pre-trained models and the observed trends in parameter updates across languages are influenced by factors such as model architecture, training procedures, data proportions, and even the order in which the model encounters training data. As a result, the specific layers we identified for each model size may differ when tested on other LLMs. Additionally, our observations suggest that certain languages are undertrained in Llama models. However, due to the lack of publicly available information on training data and procedures, we cannot make definitive claims regarding language-specific training discrepancies.

Another limitation is that our study focuses on languages that mostly come from the same language family, and are relatively close to each other. As a result these languages exhibit significant token overlap, facilitating cross-lingual transfer. The models we evaluated were also trained on a limited set of languages with similar characteristics. The studied languages mainly fall into the mid- or high-resource category, meaning our findings may not generalize to massively multilingual models trained on a more diverse set of languages. 

\subsection{Ethical Statement}
This research investigates language adherence in multilingual large language models and proposes partial training methods for efficient adaptation. Our work aims to enhance linguistic fairness and accessibility by mitigating biases that favor high-resource languages. We acknowledge that training data composition and fine-tuning decisions can introduce unintended biases, which may disproportionately affect underrepresented languages. While our findings contribute to more equitable multilingual model adaptation, they are limited to languages present in the model's pretraining data and may not generalize to unseen languages. We encourage further work to assess our method’s applicability to a broader set of languages, particularly low-resource and non-Indo-European languages.

This study does not involve human subjects, personal data, or user interactions, and we adhere to ethical guidelines for computational research. Our experiments were conducted using publicly available models and datasets, ensuring transparency and reproducibility.

\subsection{Implementation details} \label{sec:implementation-detail}
We experimented with dropout rates of 0.1 and 0.05, and learning rates of 5e-5, 1e-5, 5e-6, 1e-6, 5e-7, and 1e-7 for training on the English CBQA task. The best setting (dropout = 0.1, learning rate = 5e-6) was selected based on the minimum validation loss. These hyperparameters were used consistently across all languages and models throughout the paper.

For all training runs in our experiments, we used the hyperparameters listed in \autoref{tab:training-hyperparameter}. All experiments were conducted with a fixed random seed of 42. We implemented our models using Transformers 4.46.3 and Torch 2.5.1, with Accelerate 1.1.0 and DeepSpeed 0.16.1 for multi-GPU training. All experiments were run on NVIDIA RTX A6000 GPUs, with all experiments taking approximately 48 hours on eight GPUs. 
\begin{table}[h]
    \centering
    \caption{Training hyperparameters}
    \label{tab:training-hyperparameter}
    \begin{tabular}{lc}
    \toprule
        \textbf{Parameter} & \textbf{value} \\ \midrule
        num\_epochs & 3 \\
        save\_steps & 100 \\
        eval\_steps & 100 \\
        logging\_steps & 100 \\
        batch\_size & 64 \\
        gradient\_accumulation & 1 \\
        weight\_decay & 0.01 \\
        bf16 & True \\
        \bottomrule
    \end{tabular}
    
\end{table}

\subsection{Language specific knowledge}
\begin{figure*}
    \centering
    \includegraphics[width=0.7\textwidth]{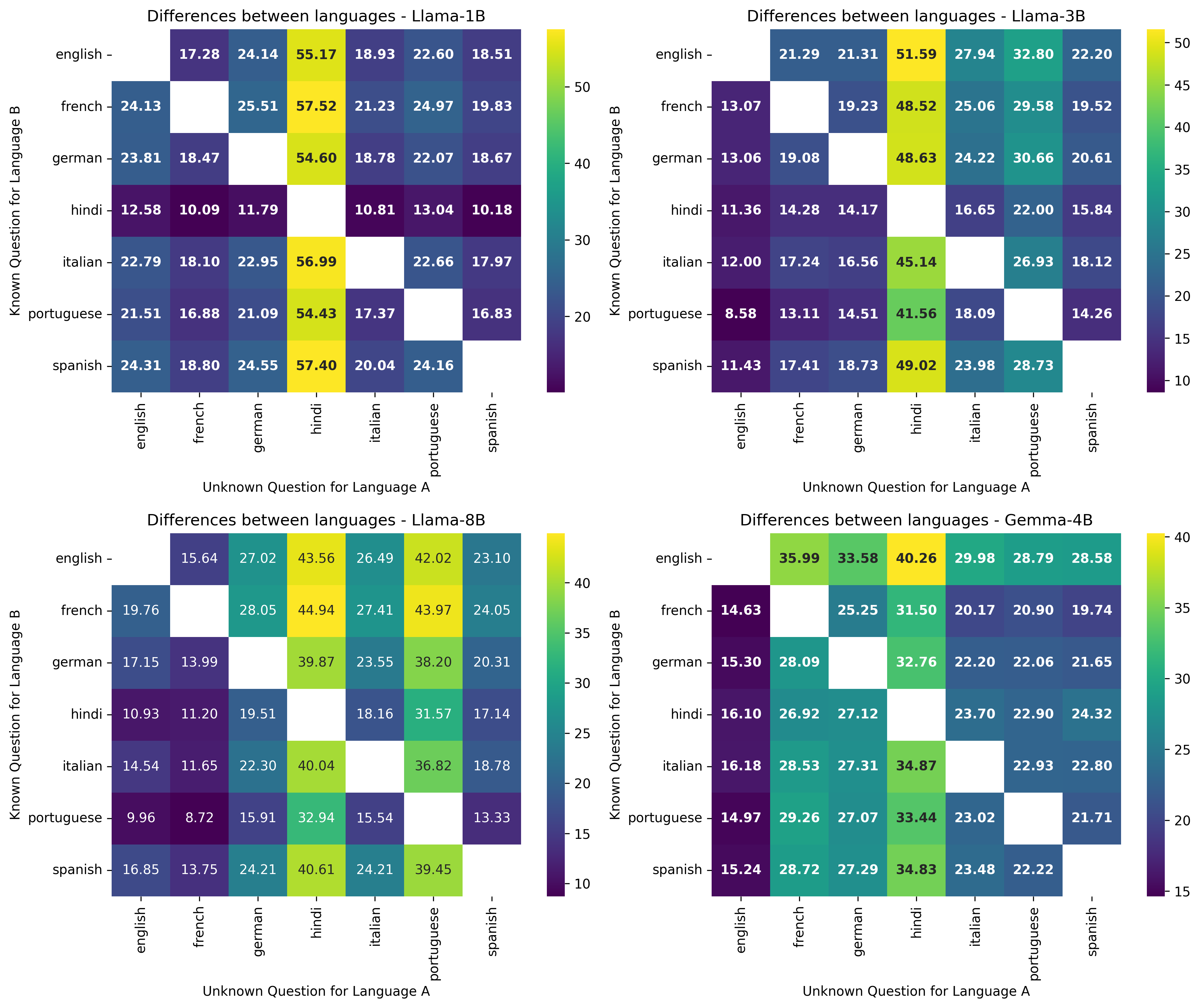}
    \caption{Difference in known knowledge between each pair of languages across different model sizes.}
    \label{fig:correct-diff}
\end{figure*}

Beyond measuring similarities between languages using the Jaccard Index, we also analyze differences by identifying answers that are known in language A but unknown in language B. This allows us to examine the distribution of languages within the 40\% of answers that are not correctly predicted by both languages. The results, presented in \autoref{fig:correct-diff}, reveal an almost symmetrical distribution of known and unknown answers across most language pairs. However, notable deviations occur for languages with significantly lower overall accuracy. Specifically, Hindi shows a greater disparity in the Llama-3.2 models, while both Hindi and Portuguese exhibit this trend in the Llama-8B model.

\subsection{Parameter update} \label{appendix:heatmap}
\begin{figure*}
    \centering
    \includegraphics[width=\linewidth]{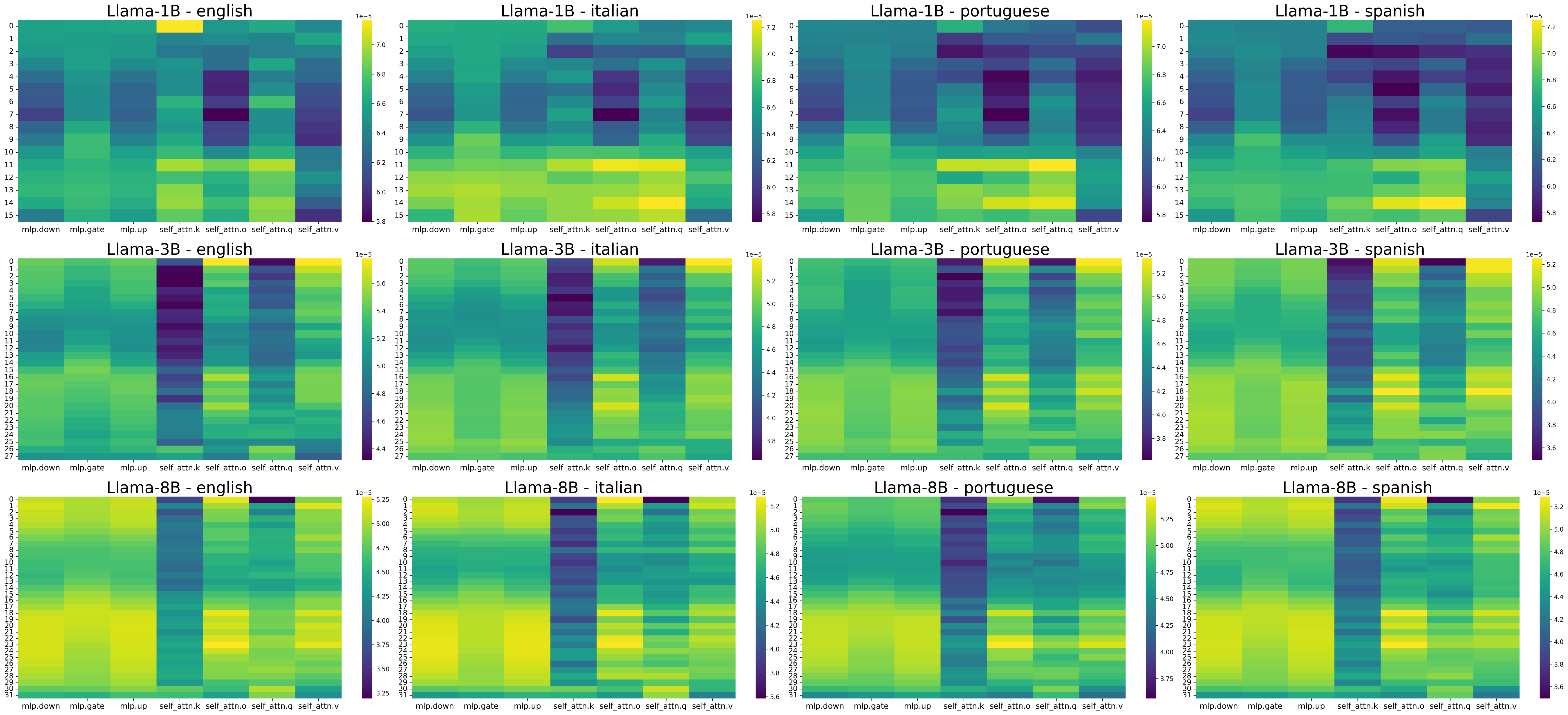}
    \caption{Average magnitude of difference between pretrained and monolingually fine-tuned models for Llama-1B, Llama-3B, and Llama-8B.}
    \label{fig:model-diff-appendix}
\end{figure*}

Due to space constraints, the main text presents results for only four languages. However, the analysis of model updates for Italian, Spanish, and Portuguese follows similar trends and can be found in \autoref{fig:model-diff-appendix}. These additional results confirm the patterns observed in other languages, reinforcing our findings on language-specific parameter updates.

\subsection{Partial Training} \label{sec:partial-appendix}
\begin{figure*}[h]
    \centering
    \includegraphics[width=\linewidth]{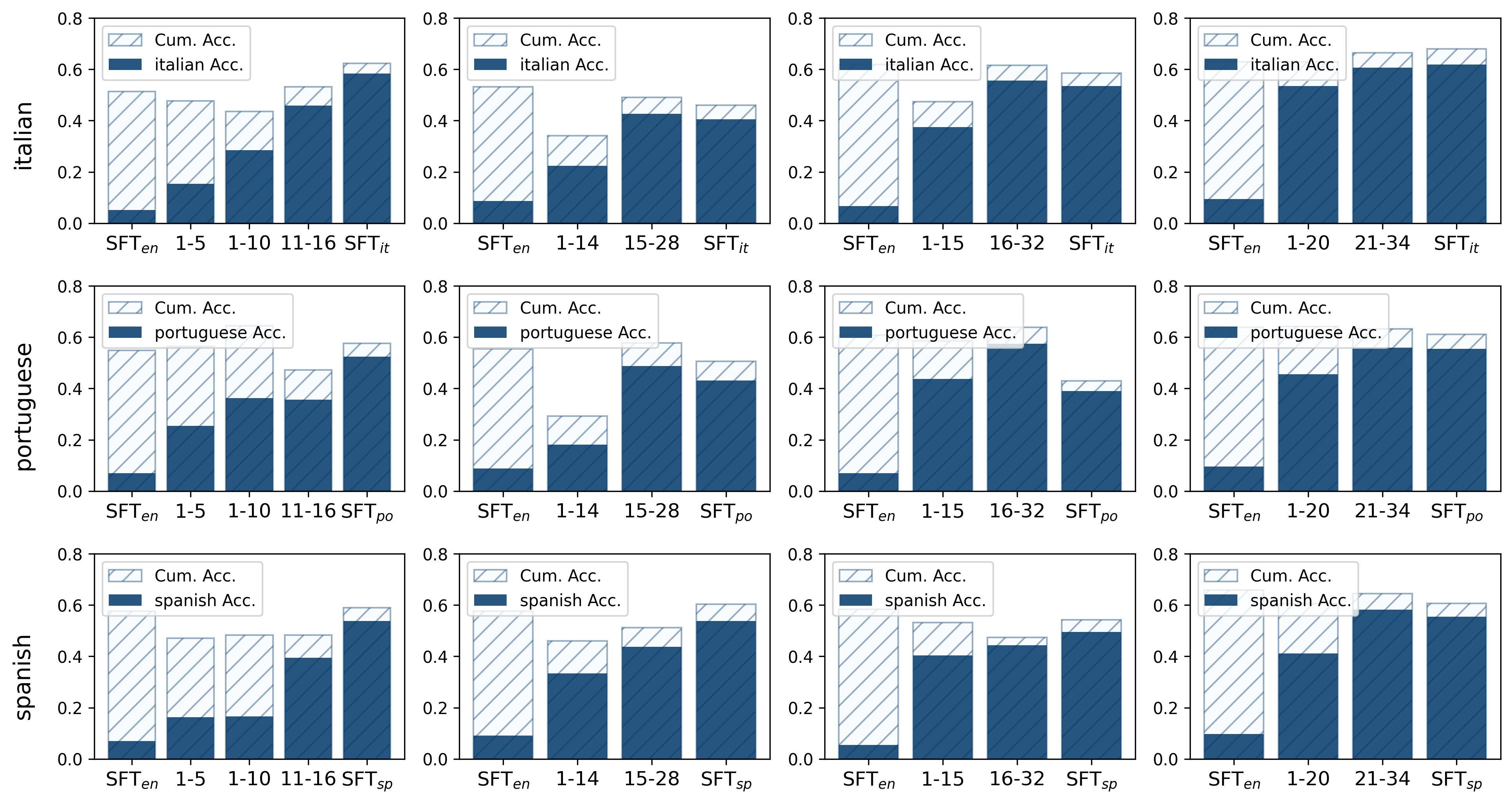}
    \caption{Cumulative accuracy and \( L_i \) accuracy on \textit{en-tuned} (\( SFT_{en} \)) and \( L_i \)-tuned models (\( SFT_{L_i} \)), along with partially trained models, across all Llama model sizes.}
    \label{fig:partial-training-appendix}
\end{figure*}

Due to space limitations, the results of partial training on Italian, Portuguese, and Spanish are provided in \autoref{fig:partial-training-appendix}. Additionally, the CoCo-CoLa ratios for both partially trained and fully trained models are shown in \autoref{tab:coca-ratio-partial-1b} for Llama-1B, \autoref{tab:coca-ratio-partial-3b} for Llama-3B, and \autoref{tab:coca-ratio-partial-8b} for Llama-8B. These comparisons highlight the consistently superior CoCo-CoLa ratio in the partial training of final layers.

  \begin{table}[h]
    \centering
    \caption{CoCo-CoLa Ratios (\%) for different languages across finetuned Llama-3.2-1B models.}
    \label{tab:coca-ratio-partial-1b}
    \begin{tabular}{lccccc}
        \toprule
        Language & SFT$_{en}$ & 1-5 & 1-10 & 11-16 & SFT$_{L_i}$ \\
        \midrule
        French      & 13.47  & 44.63  & 50.22  & 78.72  & 88.58  \\
        German      & 10.50  & 25.12  & 31.77  & 76.66  & 91.02  \\
        Hindi       & 13.28  & 56.82  & 58.49  & 92.73  & 90.79  \\
        Italian     & 10.00  & 32.12  & 65.17  & 86.18  & 93.60  \\
        Portuguese  & 12.73  & 45.18  & 56.33  & 75.43  & 91.07  \\
        Spanish     & 12.01  & 34.61  & 34.41  & 81.66  & 91.24  \\
        \bottomrule
    \end{tabular}
    
\end{table}

  \begin{table}[h]
    \centering
    \caption{CoCo-CoLa Ratios (\%) for different languages across finetuned Llama-3.2-3B models.}
    \label{tab:coca-ratio-partial-3b}
    \begin{tabular}{lcccc}
        \toprule
        Language & SFT$_{en}$ & 1-14 & 14-27 & SFT$_{L_i}$ \\
        \midrule
        French  & 14.73  & 54.64  & 81.18  & 89.45  \\
        German  & 19.64  & 71.40  & 86.04  & 89.26  \\
        Hindi  & 10.04  & 26.40  & 88.41  & 77.47  \\
        Italian  & 16.29  & 65.45  & 86.91  & 87.91  \\
        Portuguese  & 15.99  & 61.76  & 84.45  & 85.10  \\
        Spanish  & 15.84  & 72.38  & 85.50  & 89.18  \\
        \bottomrule
    \end{tabular}
    
\end{table}

  \begin{table}[h]
    \centering
    \caption{CoCo-CoLa Ratios (\%) for different languages across finetuned Llama-3.1-8B models.}
    \label{tab:coca-ratio-partial-8b}
    \begin{tabular}{lcccc}
        \toprule
        Language & SFT$_{en}$ & 1-15 & 16-31 & SFT$_{L_i}$ \\
        \midrule
        French  & 12.32  & 78.93  & 87.77  & 87.54  \\
        German  & 11.03  & 81.91  & 88.69  & 87.21  \\
        Hindi  & 10.74  & 67.08  & 96.06  & 90.68  \\
        Italian  & 10.90  & 78.92  & 90.28  & 91.35  \\
        Portuguese  & 11.49  & 74.68  & 90.11  & 90.73  \\
        Spanish  &  9.40  & 75.82  & 93.55  & 91.35  \\
        \bottomrule
    \end{tabular}
    
\end{table}

  \begin{table}[h]
    \centering
    \caption{CoCo-CoLa Ratios (\%) for different languages across finetuned Gemma-3-4B models.}
    \label{tab:coca-ratio-partial-gemma}
    \begin{tabular}{lcccccc}
        \toprule
        Language & SFT$_{en}$ & 1-20 & 21-34 & SFT$_{L_i}$ \\
        \midrule
        French      & 19.22 & 64.26 & 89.99 & 90.14 \\
        German      & 15.23 & 86.70 & 93.03 & 92.64 \\
        Hindi       &  9.39 & 92.74 & 96.30 & 97.19 \\
        Italian     & 14.84 & 85.03 & 91.20 & 91.08 \\
        Portuguese  & 14.98 & 70.93 & 88.40 & 90.69 \\
        Spanish     & 14.70 & 68.14 & 90.19 & 91.36 \\
        \bottomrule
    \end{tabular}
\end{table}

\afterpage{%
  \clearpage
    \section*{NeurIPS Paper Checklist}

The checklist is designed to encourage best practices for responsible machine learning research, addressing issues of reproducibility, transparency, research ethics, and societal impact. Do not remove the checklist: {\bf The papers not including the checklist will be desk rejected.} The checklist should follow the references and follow the (optional) supplemental material.  The checklist does NOT count towards the page
limit. 

Please read the checklist guidelines carefully for information on how to answer these questions. For each question in the checklist:
\begin{itemize}
    \item You should answer \answerYes{}, \answerNo{}, or \answerNA{}.
    \item \answerNA{} means either that the question is Not Applicable for that particular paper or the relevant information is Not Available.
    \item Please provide a short (1–2 sentence) justification right after your answer (even for NA). 
\end{itemize}

{\bf The checklist answers are an integral part of your paper submission.} They are visible to the reviewers, area chairs, senior area chairs, and ethics reviewers. You will be asked to also include it (after eventual revisions) with the final version of your paper, and its final version will be published with the paper.

The reviewers of your paper will be asked to use the checklist as one of the factors in their evaluation. While "\answerYes{}" is generally preferable to "\answerNo{}", it is perfectly acceptable to answer "\answerNo{}" provided a proper justification is given (e.g., "error bars are not reported because it would be too computationally expensive" or "we were unable to find the license for the dataset we used"). In general, answering "\answerNo{}" or "\answerNA{}" is not grounds for rejection. While the questions are phrased in a binary way, we acknowledge that the true answer is often more nuanced, so please just use your best judgment and write a justification to elaborate. All supporting evidence can appear either in the main paper or the supplemental material, provided in appendix. If you answer \answerYes{} to a question, in the justification please point to the section(s) where related material for the question can be found.

IMPORTANT, please:
\begin{itemize}
    \item {\bf Delete this instruction block, but keep the section heading ``NeurIPS Paper Checklist"},
    \item  {\bf Keep the checklist subsection headings, questions/answers and guidelines below.}
    \item {\bf Do not modify the questions and only use the provided macros for your answers}.
\end{itemize}


\begin{enumerate}

\item {\bf Claims}
    \item[] Question: Do the main claims made in the abstract and introduction accurately reflect the paper's contributions and scope?
    \item[] Answer: \answerYes{} 
    \item[] Justification: Each claim introduced in the abstract and introduction is explicitly formulated as a hypothesis in Section 4 and then directly evaluated in our experiments. 

\item {\bf Limitations}
    \item[] Question: Does the paper discuss the limitations of the work performed by the authors?
    \item[] Answer: \answerYes{} 
    \item[] Justification: The paper includes a limitations subsection in the appendix that discusses the experimental limitations and how they may affect the generalization of the work to different settings or models. 

\item {\bf Theory assumptions and proofs}
    \item[] Question: For each theoretical result, does the paper provide the full set of assumptions and a complete (and correct) proof?
    \item[] Answer: \answerNA{} 
    \item[] Justification: The paper does not include theoretical assumptions. 

    \item {\bf Experimental result reproducibility}
    \item[] Question: Does the paper fully disclose all the information needed to reproduce the main experimental results of the paper to the extent that it affects the main claims and/or conclusions of the paper (regardless of whether the code and data are provided or not)?
    \item[] Answer: \answerYes{}{} 
    \item[] Justification: All modified hyperparameters are listed in the appendix, along with the versions of all implementation tools. The dataset used is publicly available, and all modifications to it are disclosed in the main text. Criteria for selecting the best hyperparameters and models are also explained in the appendix. 

\item {\bf Open access to data and code}
    \item[] Question: Does the paper provide open access to the data and code, with sufficient instructions to faithfully reproduce the main experimental results, as described in supplemental material?
    \item[] Answer: \answerYes{} 
    \item[] Justification: The dataset used is publicly available and the code for all experiments, is included as supplementary material and will be accessible through GitHub. 

\item {\bf Experimental setting/details}
    \item[] Question: Does the paper specify all the training and test details (e.g., data splits, hyperparameters, how they were chosen, type of optimizer, etc.) necessary to understand the results?
    \item[] Answer: \answerYes{} 
    \item[] Justification: All training and test details are provided in the appendix. 

\item {\bf Experiment statistical significance}
    \item[] Question: Does the paper report error bars suitably and correctly defined or other appropriate information about the statistical significance of the experiments?
    \item[] Answer: \answerNo{}{} 
    \item[] Justification: While we were unable to repeat training with different random seeds due to limited computational resources, we note that all training involved supervised finetuning, and the random seed only affected the shuffling of training data. During testing, we used greedy decoding without sampling, minimizing the impact of randomness. Additionally, the random seed used is reported to ensure reproducibility of our results. 

\item {\bf Experiments compute resources}
    \item[] Question: For each experiment, does the paper provide sufficient information on the computer resources (type of compute workers, memory, time of execution) needed to reproduce the experiments?
    \item[] Answer: \answerYes{} 
    \item[] Justification: The appendix explicitly specifies the GPUs used for the experiments as well as the total time required to run all experiments. 
    
\item {\bf Code of ethics}
    \item[] Question: Does the research conducted in the paper conform, in every respect, with the NeurIPS Code of Ethics \url{https://neurips.cc/public/EthicsGuidelines}?
    \item[] Answer: \answerYes{} 
    \item[] Justification: We have included an Ethical Statement in the appendix that discusses the societal impact and potential harmful consequences of our research.   

\item {\bf Broader impacts}
    \item[] Question: Does the paper discuss both potential positive societal impacts and negative societal impacts of the work performed?
    \item[] Answer: \answerYes{} 
    \item[] Justification: We discuss the potential positive societal impacts in the conclusion and address possible negative impacts in the ethical statement in the appendix.
    
\item {\bf Safeguards}
    \item[] Question: Does the paper describe safeguards that have been put in place for responsible release of data or models that have a high risk for misuse (e.g., pretrained language models, image generators, or scraped datasets)?
    \item[] Answer: \answerNA{} 
    \item[] Justification: This paper and its supplementary material do not pose any high risk for misuse, so safeguards for responsible release are not necessary.

\item {\bf Licenses for existing assets}
    \item[] Question: Are the creators or original owners of assets (e.g., code, data, models), used in the paper, properly credited and are the license and terms of use explicitly mentioned and properly respected?
    \item[] Answer: \answerYes{} 
    \item[] Justification: All resources used, including models and datasets, are publicly available and have been properly cited in our experimental setup.  

\item {\bf New assets}
    \item[] Question: Are new assets introduced in the paper well documented and is the documentation provided alongside the assets?
    \item[] Answer: \answerNA{} 
    \item[] Justification: No new assets were introduced in this work.

\item {\bf Crowdsourcing and research with human subjects}
    \item[] Question: For crowdsourcing experiments and research with human subjects, does the paper include the full text of instructions given to participants and screenshots, if applicable, as well as details about compensation (if any)? 
    \item[] Answer: \answerNA{} 
    \item[] Justification:  The paper does not involve crowdsourcing or research with human subjects.

\item {\bf Institutional review board (IRB) approvals or equivalent for research with human subjects}
    \item[] Question: Does the paper describe potential risks incurred by study participants, whether such risks were disclosed to the subjects, and whether Institutional Review Board (IRB) approvals (or an equivalent approval/review based on the requirements of your country or institution) were obtained?
    \item[] Answer: \answerNA{} 
    \item[] Justification: The paper does not involve crowdsourcing or research with human subjects.

\item {\bf Declaration of LLM usage}
    \item[] Question: Does the paper describe the usage of LLMs if it is an important, original, or non-standard component of the core methods in this research? Note that if the LLM is used only for writing, editing, or formatting purposes and does not impact the core methodology, scientific rigorousness, or originality of the research, declaration is not required.
    \item[] Answer: \answerNA{} 
    \item[] Justification: We used ChatGPT solely for revising the script and fixing grammatical errors. 

\end{enumerate}
  \clearpage
}

\end{document}